# Query DAGs: A practical paradigm for implementing belief-network inference

Adnan Darwiche and Gregory Provan
Rockwell Science Center
1049 Camino Dos Rios
Thousand Oaks, CA 91360
{*darwiche, provan*}@*risc.rockwell.com*

## Abstract

We describe a new paradigm for implementing inference in belief networks, which consists of two steps: (1) compiling a belief network into an arithmetic expression called a *Query DAG* (Q-DAG); and (2) answering queries using a simple evaluation algorithm. Each non-leaf node of a Q-DAG represents a numeric operation, a number, or a symbol for evidence. Each leaf node of a Q-DAG represents the answer to a network query, that is, the probability of some event of interest. It appears that Q-DAGs can be generated using any of the standard algorithms for exact inference in belief networks — we show how they can be generated using the clustering algorithm. The time and space complexity of a Q-DAG *generation* algorithm is no worse than the time complexity of the inference algorithm on which it is based. The complexity of a Q-DAG *evaluation* algorithm is linear in the size of the Q-DAG, and such inference amounts to a standard evaluation of the arithmetic expression it represents. The main value of Q-DAGs is in reducing the software and hardware resources required to utilize belief networks in on-line, real-world applications. The proposed framework also facilitates the development of on-line inference on different software and hardware platforms due to the simplicity of the Q-DAG evaluation algorithm.

## 1  INTRODUCTION

Consider designing a car to have a self-diagnostic system that can alert the driver to a range of problems. Figure 1 shows a simplistic belief network that could provide a ranked set of diagnoses for car troubleshooting, given input from sensors hooked up to the battery, alternator, fuel-tank and oil-system.

The standard approach to building such a diagnostic system is to put this belief network, along with inference code, onto the car's computer. We have encountered a number of difficulties when using this approach to embody belief network technology in industrial applications. First, we were asked to provide the technology on multiple platforms. For some applications, the technology had to be implemented in ADA to pass certain certification procedures. In others, it had to be implemented on domain-specific hardware that only supports very primitive programming languages. Second, memory was limited to keep the cost of a unit below a certain threshold to maintain product profitability. The dilemma was the following: belief network algorithms are not trivial to implement, especially when optimization is crucial, and porting these algorithms to multiple platforms and languages would have been prohibitively expensive, time-consuming and demanding of qualified manpower.

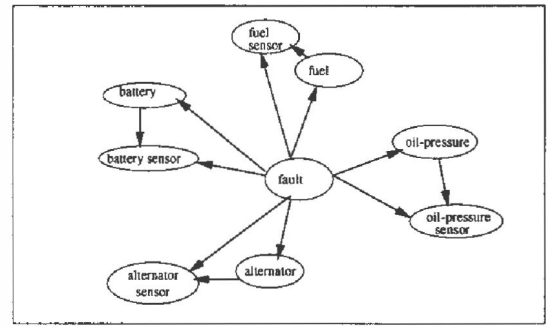

Figure 1: A simple belief network for car diagnosis.

To overcome these difficulties, we have devised a very flexible approach for implementing belief network systems, which is based on the following observation. Almost all the work performed by standard algorithms for belief networks is independent of the specific evidence gathered about variables. For example, if we run an algorithm with the battery-sensor set to *low* and then run it later with the variable set to *dead*, we find almost no algorithmic difference between the two runs. That is, the algorithm will not branch differently on any of the key decisions it makes, and the only difference between the two runs is the specific arguments to the invoked numeric operations. Therefore, one can apply a standard inference algorithm on a network with evidence being a *parameter* instead of being a specific value. The result returned by the algorithm will then be an arithmetic expression with some parameters that depend on specific evidence. This parametrized expression is what we call a Query DAG, an example of which is shown in Figure 3.

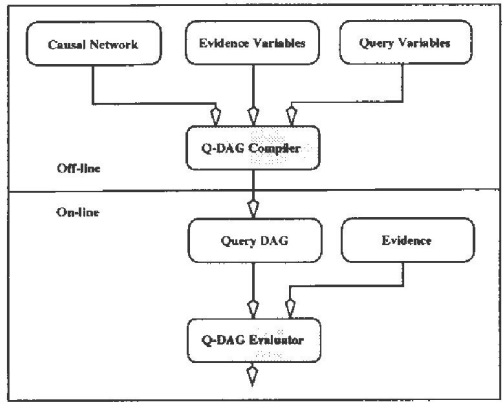

Figure 2: The proposed framework for implementing belief-network inference.

The approach we are proposing consists of two steps. First, given a belief network, a set of variables about which evidence may be collected (evidence variables), and a set of variables for which we need to compute probability distributions (query variables), a Q-DAG is compiled off-line, as shown in Figure 2. The compilation is typically done on a sophisticated software/hardware platform, using a traditional belief network inference algorithm in conjunction with the Q-DAG compilation method. This part of the process is far and away the most costly computationally. Second, an on-line system composed from the generated Q-DAG and an evaluator specific to the given platform is used to evaluate the Q-DAG. Given evidence, the parameterized arithmetic expression is evaluated in a straightforward manner using simple arithmetic operations rather than complicated belief network inference. The computational work needed to perform this on-line evaluation is so straightford that it lends itself to easy implementations on different software and hardware platforms.

This approach shares some commonality with other methods that symbolically manipulate probability expressions, like SPI [3, 5]; it differs with SPI on the objective of such manipulations and, hence, on the results obtained. SPI explicates the notion of an arithmentic expression to state that belief-network inference can be viewed as an expression-factoring operation. This allows results from optimization theory to be utilized in belief-network inference. We explicate an arithmentic expression to explicate and formalize the bound-aries between on-line and off-line inference, with the goal of identifying the minimal piece of software that is required on-line. Our results are therefore oriented towards this purpose and they include (a) a formal definition of a Q-DAG and its evaluator; (b) a method for generating Q-DAGS using any standard inference algorithm — an algorithm need not subscribe to the inference-as-factoring view to be used for Q-DAG generation; (c) computational guarantees on the size of Q-DAGs in terms of the computational guarantees of the inference algorithm used to generate them. Although the SPI framework is positioned to formulate related results, it has not been pursued in this direction.

It is important to stress the following properties of the proposed approach. First, declaring an evidence variable in the compilation process does *not* mean that evidence must be collected about that variable on-line—this is important because some evidence values, e.g., from sensors, may be lost in practice—it only means that evidence *may* be collected. Therefore, one can declare all variabes to be evidence if one wishes. Second, a variable can be declared to be both evidence and query. This allows one to perform value-of-information computations to decide whether it is worth collecting evidence about a specific variable. Third, the space complexity of a Q-DAG in terms of the number of evidence variables is no worse than the time complexity of its underlying inference algorithm; therefore, this is not a simple enumerate-all-possible-cases approach. Finally, the time and space complexity for generating a Q-DAG is no worse than the time complexity of the standard belief-network algorithm used in its generation. Therefore, if a network can be solved using a standard inference algorithm, we can construct a Q-DAG for that network.

The following section explains the concept of a Q-DAG with a concrete example and provides formal definitions. Based on this framework, it appears that any belief-network inference algorithm can be used to compile a Q-DAG as long as it meets some general conditions, a topic discussed in Section 3, which is dedicated to the generation of Q-DAGs and their computational complexity. Finally, Section 4 closes with some concluding remarks.

## 2 QUERY DAGs

Consider the belief network in Figure 3. Suppose that we typically have evidence about variable $C$ and we are interested in the probability of variable $B$. Figure 3 depicts a Q-DAG for answering such queries.

The Q-DAG has two leaf nodes corresponding to $Pr(B=ON, \mathbf{e})$ and $Pr(B=OFF, \mathbf{e})$, respectively. A root node of the form $(V, v)$ is called an *Evidence Specific Node (ESN)* and its value depends on the evidence collected about variable $V$ on-line.

The value of node $(V, v)$ is 1 if variable $V$ is instantiated to $v$ or is unknown, and 0 otherwise. Once the values of ESNs are determined, we evaluate the remaining nodes of a Q-DAG using numeric multiplication and addition. The numbers that get assigned to leaf nodes as a result of this evaluation are the answers to the queries represented by these leaf nodes.

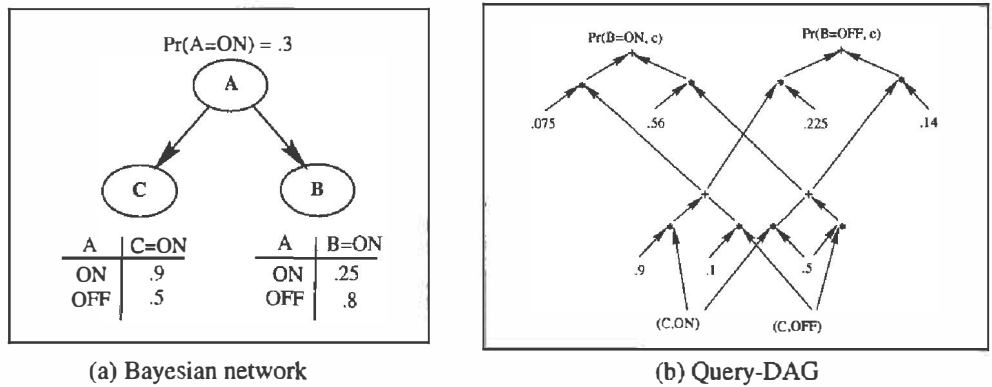

Figure 3: A probabilistic causal network (a), and the corresponding Query-DAG (b). $C$ is an evidence variable, and we are interested in the probability of Variable $B$.

For example, suppose that the evidence we have is $C = ON$. Then ESN $(C, ON)$ is evaluated to 1 and ESN $(C, OFF)$ is evaluated to 0. The Q-DAG in Figure 3(b) is then evaluated as given in Figure 4(a), leading to

$$Pr(B=ON, C=ON) = .3475,$$

and

$$Pr(B=OFF, C=ON) = .2725,$$

for which we conclude that $Pr(C=ON) = .62$.

If the evidence we have is $C = OFF$, then $(C, ON)$ evaluates to 0 and $(C, OFF)$ evaluates to 1. The Q-DAG in Figure 3(b) will then be evaluated as given in Figure 4(b), leading to

$$Pr(B=ON, C=OFF) = .2875,$$

and

$$Pr(B=OFF, C=OFF) = .0925.$$

We will use the following notation for denoting variables and their values. Variables are denoted using uppercase letters, such as $A, B, C$ and variable values are denoted by lowercase letters, such as $a, b, c$. Sets of variables are denoted by boldface uppercase letters, such as $\mathbf{A}, \mathbf{B}, \mathbf{C}$, and their instantiations are denoted by boldface lowercase letters, such as $\mathbf{a}, \mathbf{b}, \mathbf{c}$. We use $\mathbf{E}$ to denote the set of variables about which we have evidence. Therefore, we use $\mathbf{e}$ to denote an instantiation of these variables that represents evidence.

Following is the formal definition of a Q-DAG.

**Definition 1** *A Q-DAG is a tuple $(\mathcal{V}, \diamond, \mathcal{I}, \mathcal{D}, \mathcal{Z})$ where*

1. *$\mathcal{V}$ is a distinguished set of symbols (called <u>evidence variables</u>)*

2. *$\diamond$ is a symbol (called <u>unknown value</u>)*
3. *$\mathcal{I}$ maps each variable in $\mathcal{V}$ into a set of symbols (called <u>variable values</u>) different from $\diamond$.*
4. *$\mathcal{D}$ is a directed acyclic graph where*
   - *each non-root node is labeled with either $+$ or $*$*
   - *each root node is labeled with either*
     - *a number in $[0, 1]$ or*
     - *a pair $(V, v)$ where $V$ is an evidence variable and $v$ is a value*
5. *$\mathcal{Z}$ is a distinguished set of nodes in $\mathcal{D}$ (called <u>query nodes</u>)*

Evidence variables $\mathcal{V}$ correspond to network variables about which we expect to collect evidence on-line. For example, in Figure 4, $C$ is the evidence variable. Each one of these variables has a set of possible values that are captured by the function $\mathcal{I}$. For example, in Figure 4, the evidence variable $C$ has values $ON$ and $OFF$. The special value $\diamond$ is used when the value of a variable is unknown. For example, we may have a sensor variable with values "low," "medium," and "high," but then lose the sensor value when the sensor breaks. In this case, we set the sensor value to $\diamond$. Query nodes are those representing answers to user queries. For example, in Figure 4, $B$ is the query variable, and leads to query nodes $Pr(B = ON, c)$ and $Pr(B=OFF, c)$.

An important notion is that of evidence:

**Definition 2** *For a given Q-DAG $(\mathcal{V}, \diamond, \mathcal{I}, \mathcal{D}, \mathcal{Z})$, <u>evidence</u> is defined as a function $\mathcal{E}$ that maps each variable $V$ in $\mathcal{V}$ into the set of values $\mathcal{I}(V) \cup \{\diamond\}$.*

When a variable $V$ is mapped into $v \in \mathcal{I}(V)$, then evidence tells us that $V$ is instantiated to value $v$. When $V$ is mapped into $\diamond$, then evidence does not tell us anything about the value of $V$.

We can now state formally how to evaluate a Q-DAG given some evidence. But first, the following notation:

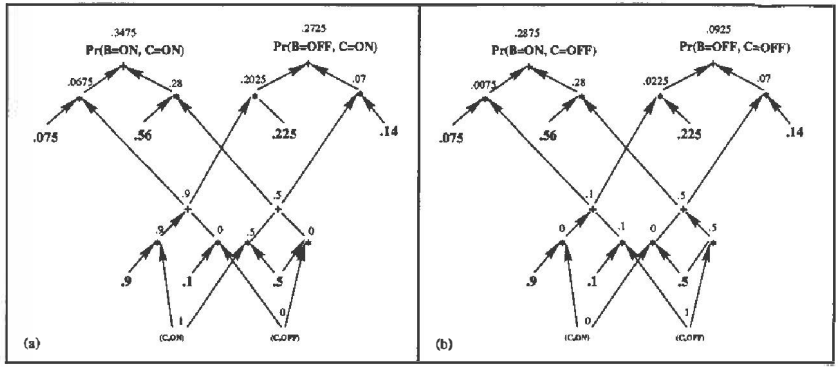

Figure 4: Evaluating the Q-DAG in Figure 3 with respect to (a) $C = ON$ and (b) $C=OFF$.

- $n(p)$ denotes a root node labeled with a number $p \in [0,1]$
- $n(V, v)$ denotes a root node labeled with $(V, v)$
- $n_1 \otimes \ldots \otimes n_i$ denotes a node labeled with * and having parents $n_1, \ldots, n_i$
- $n_1 \oplus \ldots \oplus n_i$ denotes a node labeled with + and having parents $n_1, \ldots, n_i$

The following definition tells us how to evaluate a Q-DAG by evaluating each of its nodes. It is a recursive definition according to which the value assigned to a node is a function of the values assigned to its parents. The first two cases are boundary conditions, assigning values to root nodes. The last two cases are the recursive ones.

**Definition 3** *For a Q-DAG $(\mathcal{V}, \diamond, \mathcal{I}, \mathcal{D}, \mathcal{Z})$ and evidence $\mathcal{E}$, the <u>node evaluator</u> is defined as a function $\mathcal{M}_\mathcal{E}$ that maps each node in $\mathcal{D}$ into a number $[0,1]$ such that:*

1. $\mathcal{M}_\mathcal{E}[n(p)] = p$
   *(The value of a root node labeled with a number is the number itself.)*

2. $\mathcal{M}_\mathcal{E}[n(V,v)] = \begin{cases} 1, & \text{if } \mathcal{E}(V) = v \text{ or } \mathcal{E}(V) = \diamond; \\ 0, & \text{otherwise} \end{cases}$
   *(The value of an evidence-specific node depends on the available evidence: it is 1 if $v$ is consistent with the evidence and 0 otherwise.)*

3. $\mathcal{M}_\mathcal{E}[n_1 \otimes \ldots \otimes n_i] = \mathcal{M}_\mathcal{E}(n_1) * \ldots * \mathcal{M}_\mathcal{E}(n_i)$
   *(The value of a node labeled with * is the product of the values of its parents nodes.)*

4. $\mathcal{M}_\mathcal{E}[n_1 \oplus \ldots \oplus n_i] = \mathcal{M}_\mathcal{E}(n_1) + \ldots + \mathcal{M}_\mathcal{E}(n_i)$
   *(The value of a node labeled with + is the sum of the values of its parents nodes.)*

Let us consider a few evaluations of the Q-DAG shown in Figure 3. Given evidence $\mathcal{E}(C)=ON$, we have

$$\begin{aligned}
\mathcal{M}_\mathcal{E}[n(C, ON)] &= 1, \\
\mathcal{M}_\mathcal{E}[n(C, OFF)] &= 0, \\
\mathcal{M}_\mathcal{E}[Qnode(B=ON)] &= .075 * .9 + .56 * .5 = .3475, \\
\mathcal{M}_\mathcal{E}[Qnode(B=OFF)] &= .225 * .9 + .14 * .5 = .2725,
\end{aligned}$$

meaning that $Pr(B=ON, C=ON) = .3475$ and $Pr(B=OFF, C=ON) = .2725$. Analogous computations can be done if the evidence were $\mathcal{E}(C)=OFF$.

It is also possible that evidence tells us nothing about the value of variable $C$, that is, $\mathcal{E}(C) = \diamond$. In this case, we would have

$$\begin{aligned}
\mathcal{M}_\mathcal{E}[n(C, ON)] &= 1, \\
\mathcal{M}_\mathcal{E}[n(C, OFF)] &= 1, \\
\mathcal{M}_\mathcal{E}[Qnode(B=ON)] &= .25 * .3 * (.9 + .1) + \\
& \quad .8 * .7 * (.5 + .5) = .635, \\
\mathcal{M}_\mathcal{E}[Qnode(B=OFF)] &= .75 * .3 * (.9 + .1) + \\
& \quad .2 * .7 * (.5 + .5) = .365,
\end{aligned}$$

concluding $Pr(B=ON) = .635$ and $Pr(B=OFF) = .365$.

### 2.1 Implementing a Q-DAG evaluator

A QDAG evaluator can be implemented using an event-driven, forward propagation scheme. Whenever the value of a Q-DAG node changes, one updates the value of its children, and so on, until no possible updates of values is possible. Another way to implement an evaluator is using a backward propagation scheme where one starts from a query node and updates its value by updating the values of its parent nodes. The forward propagation scheme is more selective in that it will only visit those nodes that need updating as opposed to visiting every node on which the value of query node depends. The specifics of the application will typically determine which method (or combination) will be more appropriate.

It is important that we stress the level of refinement enjoyed by the Q-DAG propagation scheme and the implications of this on the effeciency of query updates. Propagation in Q-DAGs is done at the arithmetic-operation level, which is contrasted with propagation at the message-operation level (used by many standard algorithms). Such propagation schemes are typically optimized by maintaining validity flags of messages so that only invalid messages are recomputed when new evidence arrives. This will clearly avoid some unnecessary computations but can never avoid all unnecessary ones because a message is typically too coarse for this purpose. For example, if only one entry in a message is invalid, the whole message is considered invalid. Recomputing such a message will lead to many unnecessary computations. This problem will be avoided in Q-DAG propagation since validity flags are attributed to arithmetic operations, which are the building blocks of message operations. Therefore, only the necessary arithmetic operations will be recomputed in a Q-DAG propagation scheme, leading to a more detailed level of optimization.

We also stress that the process of evaluating and updating a Q-DAG is done outside of probability theory and belief network inference. This makes the development of efficient on-line inference software accessible to a larger group of people who lack strong backgrounds in these areas.

## 2.2  The Availability of Evidence

The construction of a Q-DAG requires the identification of query and evidence variables. This may give an incorrect impression that we must know up front which variables are observed and which are not. This could be problematic in (1) applications where one may lose a sensor reading, thus changing the status of a variable from being observed to being unobserved; and (2) applications where some variables may be expensive to observe, leading to an on-line decision on whether to observe them or not (using some value-of-information computation).[1]

Both of these situations can be dealt with in a Q-DAG framework. First, as we indicated earlier, Q-DAGs allow the unknown value ⋄ to represent missing evidence which can be used to handle missing sensor readings. Second, a variable can be declared to be both query and evidence. This means that we can incorporate evidence about this variable when it is available, and also compute the probability distribution of the variable in case evidence is not available. This distribution can be used to compute the variable's value-of-information, which can then be used to decide whether to observe the variable. In the long version of this paper, we show how one can augment a Q-DAG to perform value-of-information computations [1].

## 3  GENERATING QUERY DAGs

This section shows how Q-DAGs can be generated using traditional algorithms for exact belief-network inference. In particular, we show how Q-DAGs can be generated using the clustering (join-tree, Jensen, LS) algorithm [2, 4, 6]. We also outline properties that must be satisfied by other belief network algorithms in order to adapt them for generating Q-DAGs.

## 3.1  The clustering algorithm

We provide a sketch of the clustering algorithm in this section. The reader interested in more details is referred to [4, 2, 6].

The clustering method assumes that we have constructed a cluster tree that satisfies the running intersection property (that is, a join tree) of the given belief network. This tree is a secondary structure on which the inference algorithm operates. We need the following notation to state this algorithm:

- $S_1, \ldots, S_n$ are the clusters, where each cluster corresponds to a set of variables in the original belief network.

- $\Psi_i$ is the *potential function* over cluster $S_i$, which is a mapping from instantiations of variables in $S_i$ into real numbers.

- $P_i$ is the *posterior probability distribution* over cluster $S_i$, which is a mapping from instantiations of variables in $S_i$ into real numbers.

- $M_{ij}$ is the message sent from cluster $S_i$ to cluster $S_j$, which is a mapping from instantiations of variables in $S_i \cap S_j$ into real numbers.

- **e** is the given evidence, that is, an instantiation of evidence variables **E**.

We also assume the standard multiplication and marginalization operations on potentials.

Our goal now is to compute $Pr(x, \mathbf{e})$ for each value $x$ of variable $X$ in the belief network. Given this notation, we can state the algorithm as follows:

- Potential functions are initialized using

$$\Psi_i = \prod_X Pr_X \lambda_X,$$

where $X$ is a variable whose matrix is assigned to cluster $S_i$, $Pr_X$ is the matrix for variable $X$[2], and $\lambda_X$ is the likelihood vector for variable $X$ ($\lambda_X(x)$ is 1 if $x$ is consistent with given evidence **e** and 0 otherwise).

- Posterior distributions are computed using

$$P_i = \Psi_i \prod_k M_{ki},$$

where $S_k$ are the clusters adjacent to cluster $S_i$.

- Messages are computed using

$$M_{ij} = \sum_{S_i \setminus S_j} \Psi_i \prod_{k \neq j} M_{ki},$$

where $S_k$ are the clusters adjacent to cluster $S_i$.

---

[1] Thanks to Jack Breese and Bruce D'Ambrosio for stressing this point to us.

[2] That is, a mapping from instantiations of the family of $X$ ($X$ and its parents) into conditional probabilities.

- The probability $Pr(x, \mathbf{e})$ is computed using

$$Pr(x, \mathbf{e}) = \sum_{S_i \setminus \{X\}} P_i,$$

where $S_i$ is the cluster to which $X$ belongs.

That is, to compute the probability of a variable, we must compute the posterior distribution of the cluster it belongs to. To compute the posterior distribution of a cluster, we collect messages from neighboring clusters. A message from cluster $S_i$ to $S_j$ is computed by collecting messages from all clusters adjacent to $S_i$ except for $S_j$.

### 3.2 Generating Q-DAGs

To generate Q-DAGs using the clustering method, we have to perform two steps.[3] First, we have to modify the initialization of potential functions so that the cluster tree is quantified using Q-DAG nodes instead of numeric probabilities. Second, we have to replace numeric addition and multiplication in the algorithm by analogous functions that operate on Q-DAGs:

1. Numeric multiplication $*$ is replaced by an operation $\otimes$ that takes Q-DAG nodes $n_1, \ldots, n_i$ as arguments, constructs and returns a new node $n$ with label $*$ and parents $n_1, \ldots, n_i$.

2. Numeric addition $+$ is replaced by an operation $\oplus$ that takes Q-DAG nodes $n_1, \ldots, n_i$ as arguments, constructs and returns a new node $n$ with label $+$ and parents $n_1, \ldots, n_i$.

Therefore, instead of numeric operations, we have Q-DAG-node constructors. And instead of returning a number as a computation result, we now return a Q-DAG node.

Before we state the Q-DAG clustering algorithm, realize that we now do not have evidence $\mathbf{e}$, but instead we have a set of evidence variables $\mathbf{E}$ for which we will collect evidence. Therefore, the Q-DAG algorithm will not compute an answer to a query $Pr(x, \mathbf{e})$, but instead will compute a Q-DAG node that will evaluate to $Pr(x, \mathbf{e})$ once the instantiation $\mathbf{e}$ of variables $\mathbf{E}$ is obtained.

In the following equations, potentials are mappings from variable instantiations to Q-DAG nodes (instead of numbers). For example, the matrix for variable $X$ will map each instantiation of $X$'s family into a Q-DAG node $n(p)$. Moreover, the likelihood vector for variable $X$ will map each of its instantiations $x$ into an evidence-specific node $n(X, x)$. The Q-DAG operations $\otimes$ and $\oplus$ are extended to operate on these new potentials in the same way thet $*$ and $+$ are extended in the clustering algorithm.

The new set of equations is:

---
[3]The formal proof of soundness of the proposed algorithm is given in the long version of the paper [1].

- Potential functions are initialized using

$$\Psi_i = \bigotimes_X n(Pr_X) \otimes n(\lambda_X),$$

where $X$ is a variable whose matrix is assigned to cluster $S_i$. $n(Pr_X)$ is the Q-DAG matrix for $X$, that is, a mapping from instantiations of $X$'s family into Q-DAG nodes representing conditional probabilities. $n(\lambda_X)$ is the Q-DAG likelihood vector of variable $X$. That is, $n(\lambda_X)(x) = n(X, x)$, which means that node $n(\lambda_X)(x)$ evaluates to 1 if $x$ is consistent with given evidence and 0 otherwise.

- Posterior distributions are computed using

$$P_i = \Psi_i \bigotimes_k M_{ki},$$

where $S_k$ are the clusters adjacent to cluster $S_i$.

- Messages are computed using

$$M_{ij} = \bigoplus_{S_i \setminus S_j} \Psi_i \bigotimes_{k \neq j} M_{ki},$$

where $S_k$ are the clusters adjacent to cluster $S_i$.

- The Q-DAG node for answering queries of the form $Pr(x, \mathbf{e})$ is computed using

$$Qnode(x) = \bigoplus_{S_i \setminus \{X\}} P_i,$$

where $S_i$ is the cluster to which $X$ belongs.

The only modifications we made to the clustering algorithm are (a) changing the initialization of potential functions and (b) replacing multiplication and addition with Q-DAG constructors of multiplication and addition nodes.

Given a set of evidence variables $\mathbf{E}$, and variable $X$, the Q-DAG algorithm can be used to compute the Q-DAG node $Qnode(x)$ for each value $x$. The result of such a computation will be a node that will evaluate to $Pr(x, \mathbf{e})$ for any given instantiation $\mathbf{e}$ of variables $\mathbf{E}$. We will now consider an example.

### 3.3 An example

We now show how the proposed Q-DAG algorithm can be used to generate a Q-DAG for the belief network in Figure 3(a).

We have only one evidence variable in this example, $C$. And we are interested in generating a Q-DAG for answering queries about variable $B$, that is, queries of the form $Pr(b, \mathbf{e})$. Figure 5(a) shows the cluster tree for the belief network in Figure 3(a), where the tables contain the potential functions needed for the probabilistic clustering algorithm. Figure 5(b) shows the cluster tree again, but the tables contain the potential functions needed by the Q-DAG clustering algorithm. Note that the tables are filled with Q-DAG nodes instead of numbers.

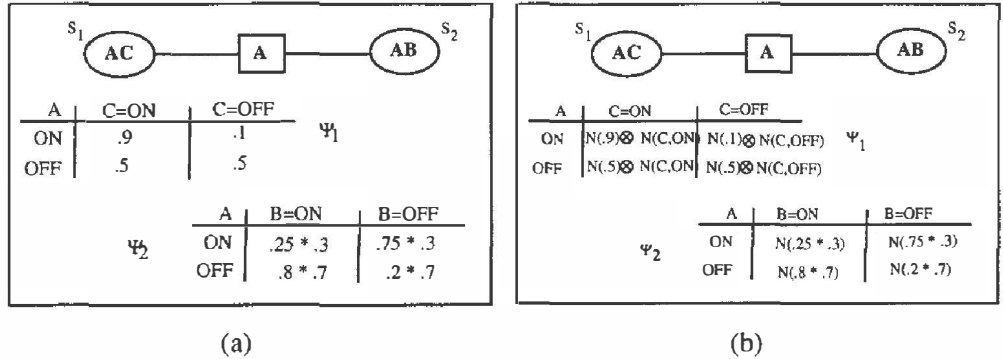

Figure 5: A cluster tree quantified with (a) numbers, and with (b) Q-DAGs.

We now apply the Q-DAG algorithm.

The message $M_{12}$ is computed by summing the potential function $\Psi_1$ over all possible values of variable $C$, i.e., $M_{12} = \bigoplus_C \Psi_1$, which leads to:

$M_{12}(A=ON)$
$= n(.9) \otimes n(C, ON) \oplus n(.1) \otimes n(C, OFF),$
$M_{12}(A=OFF)$
$= n(.5) \otimes n(C, ON) \oplus n(.5) \otimes n(C, OFF).$

The posterior distribution over cluster $S_2$, $P_2$, is computed using $P_2 = \Psi_2 \otimes M_{12}$, which leads to

$P_2(A=ON, B=ON)$
$= n(.25) \otimes n(.3) \otimes M_{12}(A=ON),$
$P_2(A=ON, B=OFF)$
$= n(.75) \otimes n(.3) \otimes M_{12}(A=ON),$
$P_2(A=OFF, B=ON)$
$= n(.8) \otimes n(.7) \otimes M_{12}(A=OFF),$
$P_2(A=OFF, B=OFF)$
$= n(.2) \otimes n(.7) \otimes M_{12}(A=OFF).$

The Q-DAG node $Qnode(b)$ for answering queries of the form $Pr(b, \mathbf{e})$ is computed by summing the posterior $P_2$ over variable $A$:

$Qnode(B=ON)$
$= P_2(A=ON, B=ON) \oplus P_2(A=OFF, B=ON),$
$Qnode(B=OFF)$
$= P_2(A=ON, B=OFF) \oplus P_2(A=OFF, B=OFF).$

Substituting and simplifying, we get

$Qnode(B=ON) =$
$n(.075) \otimes [n(.9) \otimes n(C, ON) \oplus n(.1) \otimes n(C, OFF)] \oplus$
$n(.56) \otimes [n(.5) \otimes n(C, ON) \oplus n(.5) \otimes n(C, OFF)]$

$Qnode(B=OFF) =$
$n(.225) \otimes [n(.9) \otimes n(C, ON) \oplus n(.1) \otimes n(C, OFF)] \oplus$
$n(.14) \otimes [n(.5) \otimes n(C, ON) \oplus n(.5) \otimes n(C, OFF)],$

which is the Q-DAG depicted in Figure 3.

### 3.4 Computational complexity

The computational complexity of the algorithm for generating Q-DAGs is determined by the computational complexity of the clustering algorithm. In particular, the proposed algorithm will apply a $\oplus$-operation precisely when the clustering algorithm applies an addition-operation. Similary, it will apply a $\otimes$-operation precisely when the clustering algorithm applies a multiplication-operation. Therefore, if we assume that $\oplus$ and $\otimes$ take constant time, then both algorithms have the same time complexity.

Each application of $\oplus$ or $\otimes$ operations ends up adding a new node to the Q-DAG. And this is the only way a new node can be added to the Q-DAG. Moreover, the number of parents of each added node is equal to the number of arguments that the corresponding arithmetic operation is invoked on in the clustering algorithm. Therefore, the space complexity of a Q-DAG is no worse than the time complexity of the clustering algorithm.

**Proposition 1** *The space and time complexity of generating a Q-DAG is no worse than the time complexity of the clustering algorithm.*

This also means that the space complexity of Q-DAGS in terms of the number of evidence variables is no worse than the time complexity of the clustering algorithm in those terms. Specifically, each evidence variable $E$ will only add $m$ evidence-specific nodes to the Q-DAG, where $m$ is the number of values that variable $E$ can take. This is important to stress because without this complexity guarantee it may be hard to distinguish between the proposed approach and a brute-force approach that builds a big table containing all possible instantiations of evidence variables together with their corresponding distributions of query variables.

### 3.5 Other generation algorithms

The polytree algorithm is a special case of the clustering algorithm as shown in [4]. Therefore, the polytree

210  Darwiche and Provan

algorithm can also be modified as suggested above to compute Q-DAGs. This means that cutset conditioning can be easily modified to compute Q-DAGs: for each instantiation c of the cutset C, we compute a Q-DAG node for $Pr(x, \mathbf{c})$ using the polytree algorithm and then take the $\oplus$-sum of the resulting nodes.

Most algorithms for exact inference in belief networks can be adapted to generate Q-DAGs. In general, an algorithm must satisfy a key condition to be adaptable for computing Q-DAGs as we suggested above: the algorithm should never depend on the specific evidence obtained, but should only depend on the variables about which evidence is collected. That is, whether variable $E$ is instantiated to value $v_1$ or value $v_2$ should not affect the behavior/complexity of the algorithm; only whether variable $E$ is instantiated or not should matter.

Most belief networks algorithms that we are aware of satisfy this property. The reason for this seems to be the notion of probabilistic independence on which these algorithms are based. Specifically, what is read from the topology of a belief network is a relation $I(\mathbf{X}, \mathbf{Z}, \mathbf{Y})$, stating that variables $\mathbf{X}$ and $\mathbf{Y}$ are independent given variables $\mathbf{Z}$. That is,

$$Pr(\mathbf{x}, \mathbf{y} \mid \mathbf{z}) = Pr(\mathbf{x} \mid \mathbf{z}) Pr(\mathbf{y} \mid \mathbf{z})$$

for all instantiations $\mathbf{x}, \mathbf{y}, \mathbf{z}$ of these variables. It is possible, however, for this not to hold for all instantiations of $\mathbf{z}$ but only for specific ones. The algorithms we are aware of do not take advantage of this instantiation-specific notion of independence. Therefore, they cannot attach any computational significance to the specific value to which a variable is instantiated. This property of existing algorithms is what makes them easily adaptable to the generation of Q-DAGs.

## 4  CONCLUSION

We have introduced a new paradigm for implementing belief-network inference that is oriented towards real-world, on-line applications. The proposed framework compiles a belief network into an arithmetic expression called a Query-DAG. Each non-leaf node of a Q-DAG represents a numeric operation, a number, or a symbol. Each leaf node of a Q-DAG represents the answer to a network query, that is, the probability of some event of interest. Inference on Q-DAGs is (worst-case) linear in their size and amounts to a standard evaluation of the arithmetic expressions they represent.

A most important point to stress about the work reported here is that it is *not* proposing a new algorithm for belief-network inference. What we are proposing is a paradigm for implementing belief-network inference that is orthogonal to standard inference algorithms and is engineered to meet the demands of real-world, on-line applications. This class of applications is typically demanding for the following reasons:

1. It typically requires very short response time, i.e., milliseconds.
2. It requires software to be written in specialized languages, such as ADA, C++, and assembly before it can pass certification procedures.
3. It imposes severe restrictions on the available software and hardware resources in order to keep the cost of a "unit" (such as an electromechanical device) as minimal as possible.

To address these real-world constraints, we are proposing that one compiles a belief network into a Q-DAG as shown in Figure 2 and uses a Q-DAG evaluator for on-line reasoning. This brings down the required memory to that needed for storing a Q-DAG and its evaluator. It also brings down the required software to that needed for implementing a Q-DAG evaluator, which is very simplistic as we have seen earlier.

Our proposed approach still requires a belief-network algorithm to generate a Q-DAG, but it makes the efficiency of such an algorithm less of a critical factor. For example, we show in the long version of the paper that some standard optimizations in belief-network inference, such as pruning and caching, become less critical in a Q-DAG framework. Specifically, these optimizations are subsumed by simple Q-DAG reduction techniques, such as *numeric reduction*, where a Q-DAG node $n$ and all its descendants can be replaced by a single Q-DAG node if no ESN appears among the descendants of node $n$ [1].